# Solid Waste Detection, Monitoring and Mapping in Remote Sensing Images: A Survey

Piero Fraternali[a,d], Luca Morandini[b,d] and Sergio Luis Herrera González[c,d]

[a]Email: piero.fraternali@polimi.it, ORCID: 0000-0002-6945-2625
[b]Email: luca.morandini@polimi.it, ORCID: 0009-0006-0713-3887, Corresponding author
[c]Email: sergioluis.herrera@polimi.it, ORCID: 0000-0002-8903-0622
[d]Department of Electronics, Information, and Bioengineering, Politecnico di Milano, Via Ponzio 34/5, Milan, 20133, Italy

## ARTICLE INFO

## ABSTRACT

The detection and characterization of illegal solid waste disposal sites are essential for environmental protection, particularly for mitigating pollution and health hazards. Improperly managed landfills contaminate soil and groundwater via rainwater infiltration, posing threats to both animals and humans. Traditional landfill identification approaches, such as on-site inspections, are time-consuming and expensive. Remote sensing is a cost-effective solution for the identification and monitoring of solid waste disposal sites that enables broad coverage and repeated acquisitions over time. Earth Observation (EO) satellites, equipped with an array of sensors and imaging capabilities, have been providing high-resolution data for several decades. Researchers proposed specialized techniques that leverage remote sensing imagery to perform a range of tasks such as waste site detection, dumping site monitoring, and assessment of suitable locations for new landfills. This review aims to provide a detailed illustration of the most relevant proposals for the detection and monitoring of solid waste sites by describing and comparing the approaches, the implemented techniques, and the employed data. Furthermore, since the data sources are of the utmost importance for developing an effective solid waste detection model, a comprehensive overview of the satellites and publicly available data sets is presented. Finally, this paper identifies the open issues in the state-of-the-art and discusses the relevant research directions for reducing the costs and improving the effectiveness of novel solid waste detection methods.

## 1. Introduction

The detection of illegal solid waste disposal sites is crucial for environmental protection and can help mitigate the effects of global warming and climate change (Ackerman, 2000; Manfredi et al., 2009). Improperly managed landfills most often do not control the emission of toxins that may pollute the environment and cause health risks (Alberti, 2022; Troschinetz and Mihelcic, 2009). Rainwater may also infiltrate the inappropriately disposed waste and produce leachate that contaminates soil and groundwater causing diseases to animals and humans. Last but not least, large garbage deposits exposed to high temperatures emit hot gases and heat, which may eventually lead to subsurface fires difficult to detect and extinguish (Nazari et al., 2020). Given such risks, searching and characterizing solid waste dumping sites becomes an essential task, traditionally managed with investigations and on-site inspections, which are time-consuming and costly due to the need for specialized personnel and expensive sensing infrastructures. Aerial surveys conducted with drones help speed up the monitoring process but cover limited areas, which must be identified in advance, and require operators on the field to collect images and sensor data.

Remote sensing (RS) provides an effective and economic solution for the identification and monitoring of landfills and solid waste disposal sites. The use of satellite data enables the coverage of large regions, eliminates the need for costly infrastructures, helps focus investigations on the most suspicious sites, and allows repeated acquisitions of the same area to follow the evolution of a site over time. Earth Observation (EO) satellites have been deployed for decades and a wide selection of sensors periodically collecting images are available for research or commercial purposes. Thanks to the technological evolution, revisit periods decreased over the years and daily acquisitions of the same location are feasible nowadays. At the same time, sensor resolution increased up to the point that sub- meter resolution images are the standard in modern satellites. With the increase of publicly available satellite images, a growth in the research effort has been registered and many diverse approaches have been proposed to tackle the problem of solid waste dumping site detection. Researchers developed specialized techniques that leverage various aspects of the landfill presence (e.g., visible content, spectral signatures, vegetation stress, surface temperature) for performing different tasks such as the identification of large-scale landfills or smaller dumping sites, monitoring of existing waste sites, and assessment of suitable locations for new landfills.

Only few reviews cover the techniques and data for solid waste detection and monitoring. Singh (2019) briefly surveys RS approaches for managing the environmental problems of waste disposal but does not address methods for identifying landfills or for monitoring temporal changes. Similarly, Papale et al. (2023) review a limited set of techniques to process satellite data for landfill monitoring and concentrates on case studies. Finally, Karimi et al. (2023) systematically review the use of RS imagery and analyze the geographical distributions of the adopted satellites and techniques, but the survey is very selective in the considered works.

This review provides a detailed survey of the relevant approaches for the identification and monitoring of solid waste disposal sites and of the data sets and techniques employed for such tasks. It focuses on the detection and monitoring of waste in terrestrial environments, with emphasis on large landfills and urban dumping sites containing industrial and construction waste and litter. The discussion of related tasks such as landfill mapping and environmental impact estimation helps broaden the view on the digital technologies and methods used for waste management. Ultimately, the essential goal of the survey is to show how detection and monitoring approaches vary depending on the amount and type of waste. Larger landfills can be identified by analyzing the environmental impacts due to phenomena such as the increased surface temperature or the vegetation stress. Smaller dumpsites typically found in urban environments require high-resolution imagery to recognize the complex shapes, textures and spectral signatures of suspicious materials.

The main contributions of the survey can be summarized as follows:

1. The 50 most relevant papers presented in the literature for the detection, monitoring and mapping of solid waste from remote sensing images are identified and illustrated.
2. A detailed overview of 23 EO satellites and 4 publicly available data sets used for the

development of solid waste detection models is presented and their limitations are discussed.

3. The relevant approaches are classified by the type of technique employed and by the input data used for specific waste management tasks. The proposed techniques are grouped into 7 categories ranging from visual interpretation of EO images to multi-factor analysis of multi-modal inputs to the development of Deep Learning models for satellite image analysis.

4. Research gaps, open issues and promising research directions are identified, discussed and organized into three categories: data related issues, methodological and technical limitations, and practical application needs.

5. The analysis of the literature shows that the most frequent focus is the discussion of specific case studies rather than the study of methods capable of generalizing to multiple waste disposal scenarios and geographic contexts.

6. The most significant open issues are mainly methodological. The solid waste detection and monitoring field currently lacks a standard benchmark for assessing and comparing the existing approaches, which hinders the appraisal of the research progress and the development of methodologically sound guidelines for scientist and practitioners.

7. The future research directions are extremely exciting. The waste detection and monitoring challenge can benefit from such innovations in the Computer Vision (CV) arena as Vision Transformers (Vaswani et al., 2017) or billion-scale foundation models (Cha et al., 2023), which already obtained breakthrough achievements in the natural images domain. The deployment of future satellites providing higher spatial and temporal resolution will enable the detection and classification of finer-grained waste materials and increase the accuracy of current landfill monitoring techniques. However, the technical challenges are only part of what is needed. Capturing expert knowledge for decision making and collecting evidence with court-proof methodologies are essential to turn detection into the effective investigation and prosecution of illicit waste disposal.

The review is organized as follows: Section 2 illustrates the literature search and selection methodology. Section 3 overviews the relevant approaches and describes their focuses. Section 4 presents the satellites and data sets available for research purposes, Section 5 discusses the techniques implemented by the papers, Section 6 highlights the open issues and discusses the relevant research directions and Section 7 draws the conclusions.

## 2. Methodology

This review discusses the approaches proposed in literature for detecting and monitoring solid waste disposal sites in terrestrial environments. The relevant research has been identified by following a simplified PRISMA procedure for systematic reviews (Page et al., 2021). The search has been performed on the Scopus database, as recommended in Falagas et al. (2008) and includes relevant papers until the end of 2023. The search phrases have been formed as follows:

```
<search> :- <source> AND <task> AND <subject>
<source> :- satellite images | satellite data |
            remote sensing | geographic information system
<task> :- detection | identification
<subject> :- waste | illegal dumping sites |
             illegal landfills | waste disposal sites
```

The output of the search was manually examined to retain only articles from journals, conferences, and workshops. The initial corpus composed of 1235 papers has been reduced to 890 proposals by removing duplicates. Then, it has been expanded through snowballing by adding 132 related studies citing or cited by the works, obtaining 1022 papers. Next, studies unrelated to the focus of the survey have been removed, by checking the title, keywords, and abstract of each contribution. Only contributions written in English and published with open access have been considered. Many articles related to other RS tasks or focused on out-of-scope waste categories (e.g. floating marine litter, mining waste, water pollutants, wastewater, heavy metals or hydrocarbon waste) were excluded. The reduced corpus contained 81 contributions and a final eligibility filter has been applied by reading the full-text articles. This final step yielded the 50 works considered in this survey.

## 3. Approaches to solid waste detection and monitoring in RS data

Table 1 lists the 50 analyzed papers with the names of the authors, the publication year, the countries of interest, and the focus of the research. The reported countries refer to the locations of the data used for the research.

*Table 1: The relevant proposals with their respective focus, listed by publication year.*

| Number | Reference | Year | Countries | Research focus |
|--------|-----------|------|-----------|----------------|
| 1 | Lyon | 1987 | USA | Large-scale landfill detection |
| 2 | Salleh and Tsudagawa | 2002 | Japan | Urban waste dump detection |
| 3 | Notarnicola et al. | 2004 | Italy | Large-scale landfill detection |
| 4 | Silvestri and Omri | 2008 | Italy | Large-scale landfill detection |
| 5 | Biotto et al. | 2009 | Italy | Illegal dumping risk mapping |
| 6 | Viezzoli et al. | 2009 | Italy | Large-scale landfill detection |
| 7 | Yonezawa | 2009 | Japan | Multi-temporal landfill analysis |
| 8 | Shaker and Yan | 2010 | Canada | Landfill environmental impact |
| 9 | Shaker et al. | 2011 | Canada | Multi-temporal landfill analysis |
| 10 | Bilotta et al. | 2012 | Italy | Large-scale landfill detection |
| 11 | Cadau et al. | 2013 | Italy | Multi-temporal landfill analysis |
| 12 | Alexakis and Sarris | 2014 | Greece | Landfill location assessment |
| 13 | Beaumont et al. | 2014 | Belgium | Large-scale landfill detection |
| 14 | Jordá-Borrell et al. | 2014 | Spain | Illegal dumping risk mapping |
| 15 | Yan et al. | 2014 | Canada | Multi-temporal landfill analysis |
| 16 | Lucendo-Monedero et al. | 2015 | Spain | Illegal dumping risk mapping |

| 17 | Agapiou et al. | 2016 | Greece | Urban waste dump detection |
|----|----|------|--------|----|
| 18 | Di Fiore et al. | 2017 | Italy | Large-scale landfill detection |
| 19 | Manzo et al. | 2017 | Italy | Landfill environmental impact |
| 20 | Richter et al. | 2017 | Russia | Landfill environmental impact |
| 21 | Seror and Portnov | 2018 | Israel | Illegal dumping risk mapping |
| 22 | Abd-El Monsef and Smith | 2019 | Egypt | Landfill location assessment |
| 23 | Gill et al. | 2019 | Kuwait | Large-scale landfill detection |
| 24 | Quesada-Ruiz et al. | 2019 | Spain | Illegal dumping risk mapping |
| 25 | Vambol et al. | 2019 | Ukraine | Large-scale landfill detection |
| 26 | Faizi et al. | 2020 | Pakistan | Urban waste dump detection |
| 27 | Fazzo et al. | 2020 | Italy | Landfill environmental impact |
| 28 | Nazari et al. | 2020 | USA | Landfill environmental impact |
| 29 | Chen et al. | 2021 | China | Urban waste dump detection |
| 30 | Devesa and Brust | 2021 | Argentina | Large-scale landfill detection |
| 31 | Parrilli et al. | 2021 | Italy | Urban waste dump detection |
| 32 | Torres and Fraternali | 2021 | Italy | Urban waste dump detection |
| 33 | Abou El-Magd et al. | 2022 | Egypt | Landfill environmental impact |
| 34 | Aslam et al. | 2022 | Pakistan | Landfill location assessment |
| 35 | Azmi et al. | 2022 | Malaysia | Urban waste dump detection |
| 36 | Didelija et al. | 2022 | Bosnia-Herzegovina | Urban waste dump detection |
| 37 | Karimi and Ng | 2022 | Canada | Illegal dumping risk mapping |
| 38 | Karimi et al. | 2022 | Canada | Illegal dumping risk mapping |
| 39 | Lavender | 2022 | Bosnia-Herzegovina, India, Indonesia, Kuwait, Spain | Large-scale landfill detection |
| 40 | Li et al. | 2022 | China | Urban waste dump detection |
| 41 | Rajkumar et al. | 2022 | Asia, Europe, South America | Large-scale landfill detection |
| 42 | Yailymova et al. | 2022 | Ukraine | Multi-temporal landfill analysis |
| 43 | Yang et al. | 2022 | China | Urban waste dump detection |
| 44 | Ali et al. | 2023 | Pakistan | Landfill location assessment |
| 45 | Kruse et al. | 2023 | Asia | Large-scale landfill detection |
| 46 | Révolo-Acevedo et al. | 2023 | Peru | Landfill location assessment |
| 47 | Sun et al. | 2023 | Bangladesh, India, Congo, Sri Lanka, Nigeria, China | Urban waste dump detection |
| 48 | Ulloa-Torrealba et al. | 2023 | Colombia | Urban waste dump detection |
| 49 | Yong et al. | 2023 | China | Large-scale landfill detection |
| 50 | Zhou et al. | 2023 | China | Urban waste dump detection |

Figure 1 shows an increasing temporal trend with a peak of published papers in 2022. The initial experiments on the use of satellite images to identify waste sites are from the early 2000s, with the first application dating back to 1987, a few years after the deployment of the first satellites offering access to EO data. The growing research trend is supported by the deployment of more powerful satellites and by the exponential gains in the accuracy of Deep Learning (DL) models, which made it easier to identify and track landfills. GIS tools have also become more sophisticated and affordable, allowing researchers to better analyze the geographic context of waste sites. Such technologies fueled the transition from expensive in-situ monitoring to innovative RS approaches. Societal factors intervened too. Rapid global urban expansion and demographic growth boosted the production of waste, which prompted authorities to invest in novel cost-effective waste detection and monitoring practices.

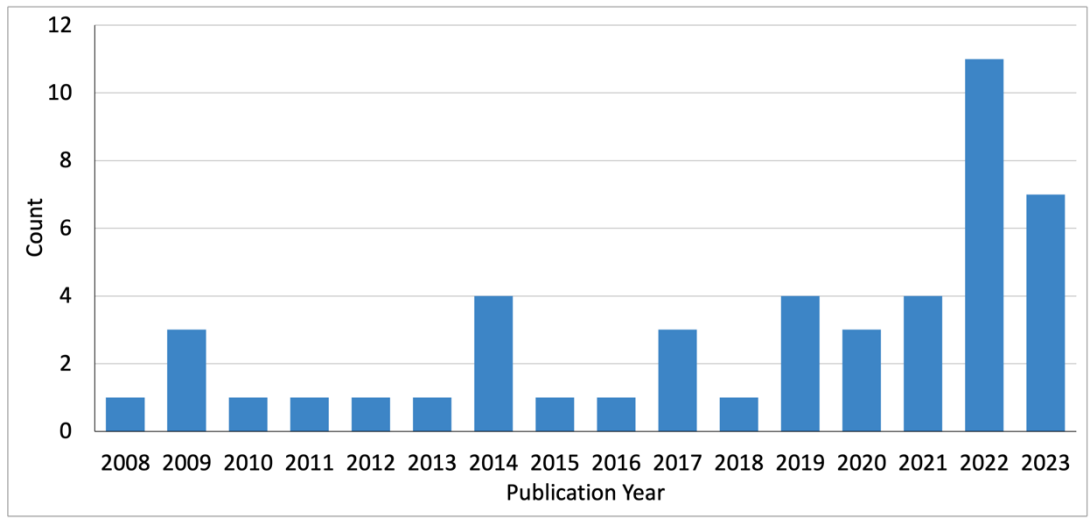

**Figure 1**: *Number of papers dedicated to solid waste detection from remote sensing data published from 2008 to 2023. An increasing trend can be identified that indicates a growing interest in this research field.*

Researchers from all over the world have addressed the solid waste disposal issue but the majority of the case studies are focused on specific regions. The mentioned countries refer to the geographical position of the data collected for training and evaluating the proposed methods, which gives a hint about where there is more concern about the problem of waste management. As Figure 2 shows, more than 75% of the relevant proposals are in Europe and Asia. However, the improper management and disposal of waste remain a global issue as demonstrated by the diversity of places and the variety of land cover (e.g. desert, urban area, forest, rural environment).

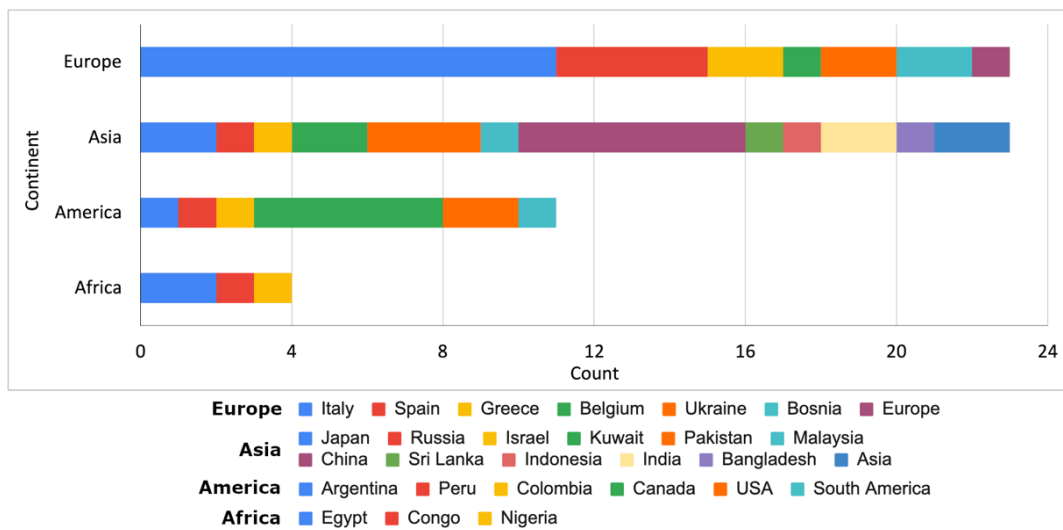

***Figure 2***: *Distribution of papers grouped by continent and by country, when available. Most works focus on detecting waste in Europe and Asia. In Europe, most research efforts are found in Italy.*

In Europe, Italy is the country that is more actively researching methods to identify illegal landfills. The country is known for the frequent discovery of unauthorized waste disposal that can harm the health of local people (Alberti, 2022).

### 3.1. Solid waste management objectives

The approaches proposed in the literature can be grouped into categories that, under the common goal of solid waste management, tackle specific problems. This review classifies the analyzed works with respect to three main objectives:

**Detection** Exploit RS data, possibly combined with other information, to analyze vast areas and identify large-scale landfills or urban dumping sites.

**Monitoring** Regularly check existing or newly identified landfills to assess their environmental impact or to detect changes over time thanks to periodical satellite acquisitions.

**Mapping** Use geographical information and RS data for determining areas with high risk of illegal dumping or for selecting suitable landfill sites.

Figure 3 shows the distribution of papers across the waste management objectives and tasks. Some approaches present a general methodology for detecting and monitoring landfills. Other works describe more specific scenarios, such as Agapiou et al. (2016), which proposes a method to identify olive oil mill waste in satellite images, or Nazari et al. (2020), which detects subsurface fires in landfills.

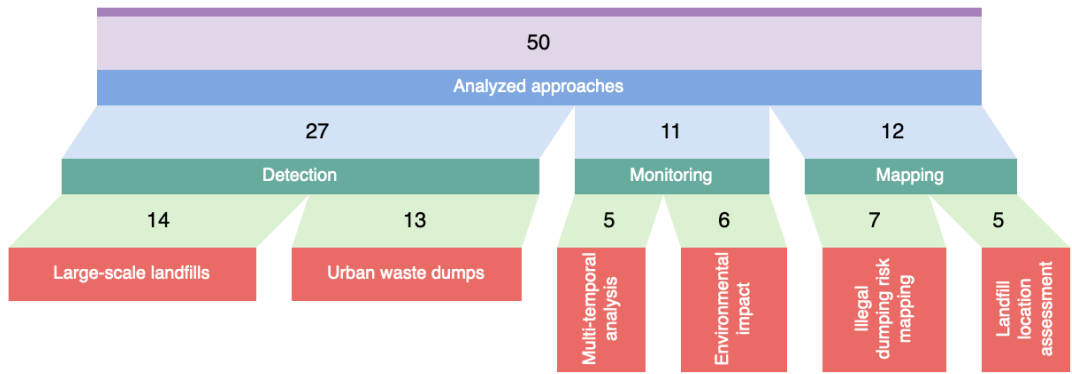

**Figure 3**: *Distribution of the analyzed approaches grouped by waste management objective and task. Most of the relevant approaches deal with the detection of large landfill sites or sparse urban solid waste.*

## 3.2. Detection tasks

**Large-scale landfills** The papers in this category aim at detecting large accumulations of solid waste and therefore apply techniques based on low resolution imagery, proving effective thanks to the large size of the landfill (Vambol et al., 2019). Some approaches analyze such signals as thermal anomalies in surface temperature (Gill et al., 2019; Beaumont et al., 2014) and stressed vegetation (Silvestri and Omri, 2008) or exploit the integration of geological and geophysical data (Di Fiore et al., 2017; Viezzoli et al., 2009). Other works use DL models to learn specific textures or spectral signatures (Devesa and Brust, 2021; Kruse et al., 2023; Rajkumar et al., 2022) relevant for detecting large landfills.

**Urban waste dumps** Some research works target small dump sites in urban environments, which can be more challenging due to the confusion induced by the surrounding background or by authorized solid waste disposal areas. In (Didelija et al., 2022), potential illegal landfill sites in urban areas are detected while (Torres and Fraternali, 2021; Salleh and Tsudagawa, 2002) focus on the identification of unauthorized waste deposits. Azmi et al., 2022 targets scattered waste in unoccupied areas and Ulloa-Torrealba et al. (2023) identifies urban waste abandoned on the streets. Some works narrow the viewpoint to specific types of solid waste: Yang et al. (2022) and Yong et al. (2023) localize demolition waste in construction sites, Agapiou et al. (2016) identifies waste produced by olive mills and Lavender (2022) detects plastic waste in marine and terrestrial environments.

## 3.3. Monitoring tasks

**Multi-temporal analysis** Some approaches exploit RS data to monitor existing landfills over time (Shaker et al., 2011; Yan et al., 2014) or to verify that waste management activities are appropriately executed (Yonezawa, 2009). In Yailymova et al. (2022) landfills are periodically monitored to analyze changes over time and similarly in Cadau et al. (2013) newly detected landfills are controlled using SAR information to identify variations in the landfill surface.

**Environmental impact** Some works focus on the threats that large-scale solid waste landfills may pose to the surrounding environment. Shaker and Yan (2010) and Manzo et al. (2017) assess the environmental impact of waste sites. Richter et al. (2017) develop a thermal model to assess the heat contamination on the environment and Nazari et al. (2020) monitor landfill subsurface fires that may happen due to high-temperature gas emissions and can lead to above-surface hazards.

### 3.4. Mapping tasks

**Illegal dumping risk mapping** Another line of research integrates RS data and geographical information to characterize the distribution of illegal dumps (Biotto et al., 2009; Quesada-Ruiz et al., 2019). Jordá-Borrell et al. (2014) and Seror and Portnov (2018) map the areas at risk of the presence of illegal landfills, while Karimi and Ng (2022) integrate RS indices to classify candidate locations. Lucendo-Monedero et al. (2015) exploit the distribution of areas at risk as a predictive model to classify suspicious sites.

**Landfill location assessment** Works in this category select the most appropriate locations to minimize environmental risk when building new landfills. In Révolo-Acevedo et al. (2023), based on remotely sensed data, the physical and biological conditions of existing landfills are analyzed to derive a map indicating the best areas for placing a new solid waste site. Similarly in Alexakis and Sarris (2014), Abd-El Monsef and Smith (2019) and Aslam et al. (2022), the optimal location is derived by combining RS vegetation indices and geographical data.

## 4. Available satellites and resources

Satellites for EO had huge improvements in the last 50 years. Today 30cm resolution with less than a day revisit time is possible. Most missions offer some level of access to research projects and a few programs provide full access to the historical data and quotas for the current ones.

### 4.1. Satellites

In this Section, we summarize the most common satellite missions used in landfill detection research and their capabilities and products.

**WorldView Series[1]:** they are owned by Maxar[2] and their data can be accessed via the European Space Agency (ESA). *WorldView-1* offers panchromatic-only images with 50cm Ground Sampling Distance (GSD) at a revisit frequency of 1.7 days. *WorldView-2* collects panchromatic images at 0.46m GSD and 8-band multispectral images at 1.8m GSD. The latter include Visible Near-Infrared (VNIR). The revisiting frequency is 1.1 days. *WorldView-3* includes panchromatic (0.31m GSD), VNIR (1.24m GSD) and 8 Shortwave Infrared (SWIR) with GSD 3.7m. The revisiting frequency is less than a day at 1m GSD (at 40º N latitude). *WorldView-4* formerly known as *GeoEye-2*, was launched in 2016 with the objective of providing large area single pass (synoptic) collection. It offered bi-directional scanning with daily revisits. The satellite stopped working in 2019, but the archive data is available.

**GeoEye-1[3]:** It is also part of the WorldView series and is optimized for large projects because it can collect over 350,000 km[2] of Very High Resolution (VHR) images per day. It provides panchromatic and VNIR multispectral imagery at 0.41m and 1.64m GSD respectively. Revisit time is 1.7 days at 1m GSD and 3 days at 0.41m GSD.

---

**Copernicus Programme[4]:** It is an EO program established by the European Commission in collaboration with ESA (Jutz and Milagro-Pérez, 2018) with most data open to citizens and institutions. The missions include:

*Sentinel-1*: launched in 2014 with a C-band Synthetic Aperture Radar (C-SAR), it provides all-weather day-and-night imagery. Four acquisition modes are used depending on the area to cover with resolutions of 5x5m, 5x20m, 20x40m, and average revisit time from less than 1 day to 3 days.

*Sentinel-2*: launched in 2015 with a multispectral sensor with 13 channels in the VNIR and SWIR spectral range, it provides 10m GSD and revisit time of 5 days around the equator and 2 to 3 days for mid-latitudes. One of its main applications is land classification.

*Sentinel-3 and other missions*: other posterior missions, such as Sentinel-3, Sentinel-5P, and Sentinel-6 feature multiple payloads but are more oriented towards the monitoring of the atmosphere and the collection of accurate topographic measurements.

**Pléiades Missions[5]:** is an environment-focused constellation owned by the French Space Agency (CNES) and operated by AirBus. Data are provided by ESA for approved projects or can be acquired from AirBus[6]. *Pléiades*: it features multispectral and VNIR sensors with 0.5m GSD for panchromatic images and 2m GSD for multispectral images. *Pléiades Neo*: provides a spatial resolution of 0.3m for panchromatic spectral bands and 1.2m for multispectral bands. Revisit time is twice a day anywhere.

**LandSat[7]:** is an EO program started by NASA in 1972 for land monitoring and applications in agriculture, cartography, geology, forestry, regional planning, surveillance, and education. Some data are freely available or can be requested through the United States Geological Survey (USGS). Two satellite missions are still active. *Landsat 8* features an Operational Land Imager (OLI) sensor, which provides panchromatic, NIR, SWIR, and a Thermal InfraRed Sensor (TIRS), which collects infrared images. It delivers 15m panchromatic and 30m multispectral GSD. *Landsat 9* features enhanced versions of the sensors (OLI-2 and TIRS-2).

**SPOT[8]:** is a high-resolution satellite system initiated by the French space agency and currently operated by Spot Image and AirBus. It supports climatology and oceanography studies and the monitoring of human activities. Data from the current mission *SPOT 6* and *SPOT 7* can be acquired from AirBus. The former provides panchromatic, and VNIR imagery at 1.5m GSD for panchromatic images and 6m GSD for multispectral ones with a revisit time of 1 to 3 days with a single satellite and 1 day with multiple satellites. The latter carries a SPOT Vegetation sensor operating in four spectral bands (blue, red, near-infrared, and middle-infrared) designed to study the spatial and temporal evolution of vegetation at multiple scales.

**COSMO-SkyMed[9]:** funded by the Italian Ministry of Research and by the Italian Ministry

---

4 https://www.copernicus.eu/en

5 https://earth.esa.int/eogateway/missions/pleiades

6 https://www.intelligence-airbusds.com/imagery/constellation/pleiades/

7 https://landsat.gsfc.nasa.gov/

8 https://earth.esa.int/eogateway/missions/spot

9 https://earth.esa.int/eogateway/missions/cosmo-skymed

of Defense, supports cartography, forest and environment protection, natural resources exploration, land management, defense and security, maritime surveillance, food and agriculture management. The current generation uses a SAR sensor with 5 modes and resolutions ranging from 0.3x0.5m to 40x6m. Data can be accessed through ESA for approved research projects and acquired from e-GEOS[10].

**ALOS[11]:** The Advanced Land Observing Satellite (ALOS), managed by the Japanese Aerospace Exploration Agency (JAXA), supports applications such as cartography, regional observation, disaster monitoring, and resource survey. The *ALOS-2* satellite operates a Phased Array L-band Synthetic Aperture Radar (PALSAR) and a Compact InfraRed Camera (CIRC), with multiple operation modes and resolutions ranging from 1x3m for 25 km² scan to 100m for 490km swath. *ALOS-1* data can be accessed through ESA and *ALOS-2* data can be acquired from PASCO[12].

**Terra[13]:** is a scientific satellite operated by NASA under the Earth Observing System (EOS) for taking atmosphere, land, and water measurements and for monitoring human activity. *Terra* carries 5 sensors, including: Advanced Spaceborne Thermal Emission and Reflection Radiometer (ASTER) creates high-resolution images of clouds, ice, water, and land with 3 sensor subsystems (SWIR, TIR, and VNIR); and Moderate-resolution Imaging Spectroradiometer (MODIS). Terra images have 15m GSD for SWIR, 30m for VNIR, and 90m for TIR with 60km swath. Data is freely available.

**Gaofen-2[14]:** is the second high-resolution EO satellite developed by China National Space Administration (CNSA) as part of CHEOS (China High-resolution Earth Observation System) program. The mission supports applications in the agriculture, disaster, resource, and environment fields. The satellite is equipped with two panchromatic sensors (0.8m GSD) and two multispectral sensors (3.2m GSD) that are combined to increase the swath width up to 45.3km. The revisit time is in the range of 4-69 days.

Other satellite missions have been deployed for medium or high resolution multispectral and hyperspectral imaging, such as *EnMAP[15]*, *Resurs-P[16]*, *PROBA-1[17]*, *GHOSt[18]*, *Aleph-1 constellation[19]*, *PRISMA[20]*. However, at the time of publication of this survey, no paper uses the data produced from such missions for solid waste detection.

Most approaches leverage multispectral images to calculate indices from a subset of bands, analyze the landfill spectral signature, or obtain high-resolution RGB images by combining channels in the visible spectrum with panchromatic images. Only a few works use data collected by SAR sensors. Lavender (2022) exploits SAR backscatter images from *Sentinel-1* to derive the surface roughness used to detect plastic waste. Cadau et al. (2013) use SAR interferometric acquisitions from *COSMO-SkyMed* to measure landfill surface variations.

Although very high-resolution imagery is available nowadays, it is important to evaluate

---

the data requirements because there is a direct relation between resolution and cost, not only for data acquisition but also for storage and processing. Finding the balance between cost, data quality (resolution), timeliness (revisit time) and the desired results is a key aspect in the design of waste detection solutions. To detect or monitor large-scale landfills through surface temperature or stressed vegetation, satellites such as *Sentinel* or *Landsat* provide sufficient resolution at affordable or no cost. In contrast, identifying smaller urban dumps requires high-resolution data, available from satellites such as *WorldView* or *Pléiades* but at a cost.

Table 2 summarises the satellites and sensor data used by the analyzed papers.

***Table 2****: List of EO satellite missions with their onboard RS payload and the respective revisit time. For each mission, references to the papers (defined in Table 1) that employ the satellite data are reported.*

| Mission | Papers numbers | Payload and resolution | Revisit time (days) |
|---------|----------------|------------------------|---------------------|
| WorldView-1 | (25) | Panchromatic (0.5m) | 1.7 |
| WorldView-2 | (17, 19, 29, 41, 50) | Panchromatic (0.46m) Multispectral (1.84m) | 1.1 |
| WorldView-3 | (41) | Panchromatic (0.31m) Multispectral (1.24m) SWIR (3.7m) | 1 |
| GeoEye-1 | (17, 25, 41) | Panchromatic (0.41m) Multispectral (1.64m) | 1.7 (at 1m GSD) 3 (at 0.41m GSD) |
| Sentinel-1 | (39) | SAR (5-40m) | 1-3 (based on latitude) |
| Sentinel-2 | (26, 30, 39, 42, 45) | Multispectral (10-60m) | 5 (at equator) 2-3 (at mid-latitudes) |
| Pléiades | (17, 19, 25, 31, 35, 36) | Panchromatic (0.5m) Multispectral (2m) | 1 |
| Landsat 4 | (20) | MSS+TM: Multispectral (57-79m) Thematic Mapper (30-120m) | 16 |
| Landsat 5 | (3, 8, 9, 13, 15, 23, 28, 33) | MSS+TM: Multispectral (57-79m) Thematic Mapper (30-120m) | 16 |
| Landsat 7 | (12, 13, 23) | ETM+: Panchromatic (15m) Multispectral (30-60m) | 16 |
| Landsat 8 | (20, 22, 26, 28, 33, 34, 38, 42, 44, 46) | OLI: PAN+MS (15-30m) TIRS: thermal (100m) | 16 |
| SPOT 4 | (12) | Panchromatic (10m) Multispectral (20m) SWIR (20m) | 26 |
| SPOT 5 | (11, 24) | Panchromatic (2.5-5m) Multispectral (10m) SWIR (20m) | 26 |
| SPOT 6 | (17, 50) | Panchromatic (1.5m) | 1-3 |

| | | Multispectral (6m) | |
|---|---|---|---|
| SPOT 7 | (35) | Panchromatic (1.5m) | 1-3 |
| | | Multispectral (6m) | |
| COSMO-SkyMed | (11, 17) | SAR: Spotlight (1m) | 5 |
| | | SAR: Stripmap (3-15m) | |
| | | SAR: ScanSAR (30-100m) | |
| ALOS | (7) | Panchromatic (2.5m) | 46 |
| | | Multispectral (10m) | |
| | | PALSAR (10-100m) | |
| Terra | (11, 13, 22, 37) | ASTER: | 4-16 |
| | | VNIR (15m) | |
| | | SWIR (30m) | |
| | | TIR (90m) | |
| IKONOS | (4, 5, 10, 25) | Panchromatic (0.82m) | 3 |
| | | Multispectral (3.28m) | |
| EO-1 | (17) | Hyperion+ALI: | 16 |
| | | Panchromatic (10m) | |
| | | Multispectral (30m) | |
| | | Hyperspectral (30m) | |
| QuickBird | (7, 17, 25) | Panchromatic (0.61-0.72m) | 1.5 |
| | | Multispectral (2.44-2.6m) | |
| RapidEye | (11) | Multispectral (6.5m) | 1 |
| Gaofen-2 | (29, 43) | Panchromatic (0.8m) | 4-69 |
| | | Multispectral (3.2m) | |

## 4.2. Datasets

In this section, we present the publicly available datasets that are specific for solid waste detection from satellite images.

**Automatic Detection of Landfill Using Deep Learning[21]:** contains the geo-locations of large landfills. Images are from Asia, Europe, and South America and sourced from WorldView-3, WorldView-2, and GeoEye-1. It is annotated with one class and includes segmentation masks.

**SWAD[22]:** comprises images collected in the Henan Province in China. It is sourced from Google Earth, WorldView-2, and SPOT. It provides annotations for one class and includes bounding boxes around the waste objects.

**Global Dumpsite Test Data[23]:** covers several large cities in Africa and Asia and is mostly sourced from Google Earth. It provides annotations for 4 classes of waste (domestic, construction, agricultural, and covered).

**AerialWaste[24]:** contains images collected from the Lombardy region in Italy and is sourced from the Italian Agriculture Development Agency (AGEA), WorldView-3, and

---

[21] https://github.com/AnupamaRajkumar/LandfillDetection_SemanticSegmentation
[22] https://www.kaggle.com/datasets/shenhaibb/swad-dataset
[23] https://www.scidb.cn/en/s/6bq2M3
[24] http://aerialwaste.org/

Google Earth. It provides binary labels (presence or absence of waste), multi-class multi-label divided into 15 solid waste types and 7 storage modes, and segmentation masks surrounding relevant waste objects.

Table 3 summarizes the characteristics of the analyzed datasets.

*Table 3: List of the publicly available RS solid waste detection datasets. For each dataset, the supported Computer Vision tasks and the statistics about the number of classes, images, and annotations are presented.*

| Data set | Year | Task | Countries | Number of classes | Number of images | Number of annotations |
|---|---|---|---|---|---|---|
| Automatic Detection of Landfill Using Deep Learning (Rajkumar et al.) | 2022 | Semantic segmentation | Serbia, Brazil, Hungary, Malta, India, Germany | 1 | 3390 | 297 |
| SWAD Dataset (Zhou et al.) | 2022 | Object detection | China | 1 | 1996 | 5562 |
| Global Dumpsite Test Data (Yin) | 2023 | Object detection | Bangladesh, India, Sri Lanka, China, Nigeria, Congo | 4 | 2219 | 2500 |
| AerialWaste (Torres and Fraternali) | 2023 | Image classification, object detection | Italy | 22 | 10434 | 3478 |

Most analyzed works do not release the dataset used for the experimentation, and in some cases, omit a detailed description of the data. This lack of transparency hinders the replication, validation, and comparison of the approaches. Furthermore, some works employ a small set of images and apply the proposed approach to specific contexts, which might not generalize to other scenarios.

The entries listed in Table 3 are a first step towards a global dataset that could be used for benchmarking. However, limitations still exist such as lack of scenario generalization or coarse annotations that do not support the classification of waste materials. Finally, all datasets offer a snapshot of the area they cover and thus do not support the evaluation of the evolution of waste dumping over time.

## 4.3. Data types employed

RS images can be used alone or in conjunction with other information. Table 4 reports the data types exploited by the analyzed works. For each paper, the employed satellites are listed along with the GIS variables and the other non-geographical information. Some works do not leverage satellite products directly but download satellite images from web archives or are focus on inputs such as GIS factors or other sources of aerial images.

**Table 4**: *Satellites, GIS data, and other data sources employed by the relevant proposal considered in the survey. Some works download archive satellite images from web archives (e.g. Google Earth) while others leverage only GIS factors or other data sources such as aerial images or geophysical measurements.*

| Reference | Year | Satellites | GIS data | Other data |
|---|---|---|---|---|
| Lyon | 1987 | - | - | Historical maps, aerial images |
| Salleh and Tsudagawa | 2002 | - | - | Aerial images |
| Notarnicola et al. | 2004 | Landsat 5 | Land cover maps | Digital orthophotos |
| Silvestri and Omri | 2008 | IKONOS | Road network | Historical aerial images, population |
| Biotto et al. | 2009 | IKONOS | Land cover maps, road network, industrial sites, location of quarries, location of authorized landfills | Population |
| Viezzoli et al. | 2009 | - | - | Ground electrical resistivity and magnetic data (SkyTEM sensor mounted on a helicopter) |
| Yonezawa | 2009 | ALOS, QuickBird | - | - |
| Shaker and Yan | 2010 | Landsat (TM) | - | - |
| Shaker et al. | 2011 | Landsat (TM) | - | Landfill gas monitoring data |
| Bilotta et al. | 2012 | IKONOS | - | Cadastral data |
| Cadau et al. | 2013 | COSMO-SkyMed, SPOT 5, RapidEye, Terra (ASTER) | - | - |
| Alexakis and Sarris | 2014 | SPOT 4, Landsat (ETM+) | Road network, land use maps, terrain morphology, distance from cities and airports, wind orientation, surface water, ... | - |
| Beaumont et al. | 2014 | Landsat 5, Landsat 7, Terra (ASTER) | Land cover maps | Aerial thermal images, Google Earth archive |
| Jordá-Borrell et al. | 2014 | - | Vegetation, altitude, land use maps, agricultural use, road network, residential areas, ... | Accessibility and visibility of illegal landfills, taxes, population, industries, urban waste, ... |
| Yan et al. | 2014 | Landsat (TM) | - | - |
| Lucendo-Monedero et al. | 2015 | - | Land use maps, slope, urban areas, road network, ... | Recycling facilities, population waste disposal frequency, industries, per capita income, ... |
| | | Solid Waste Detection in Remote Sensing Images: A Survey | | |
| Agapiou et al. | 2016 | Pléiades, SPOT 6, GeoEye-1, WorldView-2, QuickBird, COSMO-SkyMed, EO-1 | - | - |
| Di Fiore et al. | 2017 | - | - | Ground electrical resistivity and magnetic data, aerial orthophotos |
| Manzo et al. | 2017 | Pléiades, WorldView-2 | - | Aerial orthophotos, ground chemical data, field |
| Richter et al. | 2017 | Landsat 4, Landsat 8 | - | - |
| Seror and Portnov | 2018 | - | Road network, slope, forest proximity | Landfill site visibility, amount of waste |
| Abd-El Monsef and Smith | 2019 | Landsat 8, Terra (ASTER) | Waste source locations, slope, infrastructure, land use maps, road network, topography, ... | Population |
| Gill et al. | 2019 | Landsat (TM, ETM+) | - | Landfill gas measurements |
| Quesada-Ruiz et al. | 2019 | SPOT 5 | Land use maps, elevation, slope, buildings maps, cadastral data | Aerial orthophotos, population, per capita income, economic indicators, industrial indicators, waste types, ... |
| Vambol et al. | 2019 | QuickBird, WorldView-1, GeoEye-1, Pléiades, IKONOS | - | - |
| Faizi et al. | 2020 | Landsat 8, Sentinel 2 | Land cover maps | |
| Fazzo et al. | 2020 | - | Location of waste sites | Waste hazard, population |
| Nazari et al. | 2020 | Landsat 5, Landsat 8 | - | - |
| Chen et al. | 2021 | WorldView-2, Gaofen-2 | - | - |
| Devesa and Brust | 2021 | Sentinel 2 | Location of waste sites | - |
| Parrilli et al. | 2021 | Pléiades | - | - |
| Torres and Fraternali | 2021 | - | Location of waste sites | RGB aerial orthophotos |
| Abou El-Magd et al. | 2022 | Landsat 5, Landsat 8 | Land use/land cover maps, DEM data, geophysical data | Rainfall data, population |
| Aslam et al. | 2022 | Landsat 8 | Water bodies, road network, vegetation | Population, air quality, temperature |
| Azmi et al. | 2022 | Pléiades, SPOT 7 | Road network, vegetation land cover maps, residential areas | - |
| Didelija et al. | 2022 | Pléiades | Location of illegal landfill sites | - |
| Karimi and Ng | 2022 | Terra, Aura | Road network, location of waste sites | Nighttime light imagery |

Solid Waste Detection in Remote Sensing Images: A Survey

| Karimi et al. | 2022 | Landsat 8 | Road and railway network, location of waste sites | Nighttime light imagery |
|---|---|---|---|---|
| Lavender | 2022 | Sentinel 1, Sentinel 2 | - | DEM data from SRTM (Shuttle Radar Topography Mission) |
| Li et at. | 2022 | - | - | Google Earth satellite images |
| Rajkumar et al. | 2022 | GeoEye-1, WorldView-2, WorldView-3 | - | - |
| Yailymova et al. | 2022 | Landsat 8, Sentinel 2 | Location of waste sites | - |
| Yang et al. | 2022 | Gaofen-2 | - | - |
| Ali et al. | 2023 | Landsat 8 | Ground and surface water, road and railway networks, airports, land use maps | Wind direction, population |
| Kruse et al. | 2023 | Sentinel 2 | Land cover maps | - |
| Révolo-Acevedo et al. | 2023 | Landsat 8 | Land use maps, soil type | Temperatures, humidity, rainfall data |
| Sun et al. | 2023 | - | - | Google Earth satellite images |
| Ulloa-Torrealba et al. | 2023 | - | Building rooftops footprint | Multispectral aerial images |
| Yong et al. | 2023 | - | Location of construction waste landfills | Images from Esri World Imagery WMTS |
| Zhou et al. | 2023 | WorldView-2, SPOT | - | - |

## 5. Techniques

The surveyed works span over three decades and have applied a broad range of techniques, which can be grouped into homogeneous categories:

1. **Visual interpretation**: some works exploit EO images interpreted by human experts.

2. **Descriptive indices extraction and analysis**: EO multispectral images are used to extract descriptive indices by combining information from multiple spectral bands, typically about surface temperature and vegetation, which are then used for waste detection.

3. **Multi-factor analysis**: some methods exploit multi-modal inputs (e.g., GIS data and descriptive indices) as inputs to waste detection and mapping algorithms.

4. **Features extraction and classification**: some methods extract descriptive features from EO images (e.g., brightness, spectral or spatial features), which are then classified with a decision algorithm (e.g., Bayesian classifier, unsupervised classification, etc.) to detect waste accumulations.

5. **Traditional CV techniques**: CV techniques areapplied to detect waste by implementing traditional image-based tasks such as pixel-level multi-label classification, object-based classification, and segmentation.

6. **Deep Learning CV techniques**: recent works apply the advances in CV brought by Deep Learning architectures to the analysis of EO images for waste detection by addressing such tasks as image binary classification, object detection, and semantic segmentation.

7. **Other approaches**: A few works employ ad hoc procedures that do not fall in any of the preceding categories.

**Figure 4** shows the techniques for detecting, monitoring and mapping solid waste. Most methods can be used for both detecting and monitoring waste landfills, while only multi-factor analysis has been proposed for mapping areas at risk of illegal waste dumping. The following sections illustrate the proposed techniques based on the task they address.

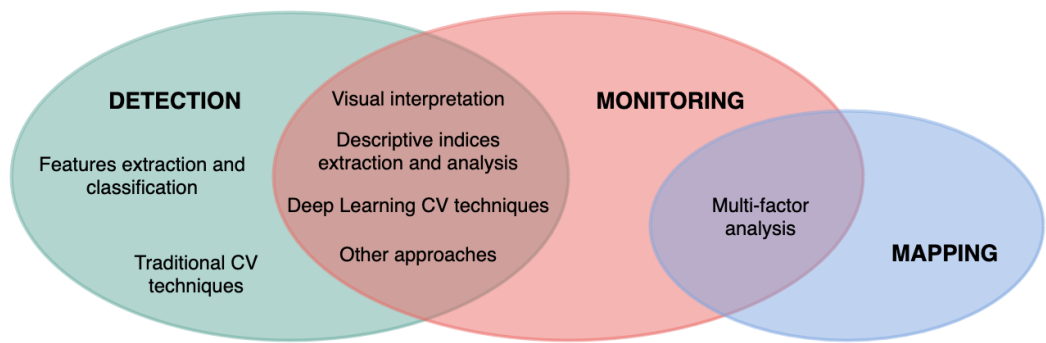

**Figure 4**: *Distribution of techniques among waste management objectives*

## 5.1. Visual interpretation

Only a few works use the manual interpretation of EO images due to the high human effort required, which hinders application at scale. Furthermore, this approach can introduce errors caused by photointepreters' biases.

**Detection** This approach was employed in the pioneering work by Lyon (1987), when automatic image processing was not common. Another work inspects optical EO images to locate areas at risk, such as unoccupied land, sick vegetation, uneven terrain texture, or darker waters (Azmi et al., 2022).

**Monitoring** Yonezawa (2009) inspects pansharpened imagery visually to ensure appropriate operation of known dumping sites and to identify potential illegal sites.

## 5.2. Descriptive indices extraction and analysis

A widely adopted approach for detecting solid waste employs indices extracted from multispectral images. The most common ones are LST (Land Surface Temperature), for estimating the surface temperature, and vegetation indices, such as NDVI (Normalized Difference Vegetation Index) and SAVI (Soil-Adjusted Vegetation Index), to identify unhealthy vegetation, which is a good indicator of landfill presence.

**Detection** The LST is one of the most used indicators for detecting large waste disposal sites (Beaumont et al., 2014), especially during summer, when the heat flux emitted by the decomposition process is more pronounced (Gill et al., 2019). Also the NDVI index is employed to identify vegetation stress induced by large disposal sites (Silvestri and Omri, 2008). Finally, Agapiou et al. (2016) introduce the domain-specific index named OOMWI (Olive Oil Mill Waste Index) for detecting dangerous olive oil mill waste.

**Monitoring** Cadau et al. (2013) use thermal anomalies in the LST and the proposed domain-specific index DDI (Dump Detection Index) for identifying new landfills. Then, variations of the landfill surface are monitored using SAR data. Yan et al. (2014) and Richter et al. (2017) use the LST to monitor the temperature difference between the landfill and the surrounding area while Shaker and Yan (2010), Shaker et al. (2011) and Manzo et al. (2017) combine the LST with other spectral indices such as NDVI, SAVI, GEMI (Global Environmental Monitoring Index) with ground-based measurements to monitor the environmental impact. Nazari et al. (2020) also employ the LST to identify and monitor subsurface fires within landfills.

## 5.3. Multi-factor analysis

Geographical variables and information derived from RS imagery are combined in several waste management tasks.

**Monitoring** Fazzo et al. (2020) proposes an indicator to monitor illegal landfill hazard based on GIS variables such as the site extension and distance to population. The aim is to estimate the human health risk of populations nearby waste sites.

**Mapping** Another common technique for predicting the presence of landfills consists of defining a probability map that estimates the locations at risk. The spatial probability distribution is derived from the combination of multiple variables associated with known dumping sites. The considered variables can be georeferenced information (e.g. road network, land cover maps) (Jordá-Borrell et al., 2014; Biotto et al., 2009; Seror and Portnov, 2018) or other types of data such as population distribution (Abd-El Monsef and Smith, 2019; Aslam et al., 2022), economic and industrial indicators (Quesada-Ruiz et al., 2019), frequency of waste disposal (Lucendo-Monedero et al., 2015), distance from cities and airports (Alexakis and Sarris, 2014), rainfall data (Révolo-Acevedo et al., 2023), wind direction (Ali et al., 2023), or nighttime light imagery used as a proxy of population density (Karimi and Ng, 2022; Karimi et al., 2022). Some works also include vegetation indices (Biotto et al., 2009) or the LST distribution (Karimi and Ng, 2022; Karimi et al., 2022; Révolo-Acevedo et al., 2023). The combination of vegetation indices and GIS data is exploited for selecting the most suitable location for building a new landfill (Alexakis and Sarris, 2014) to mitigate the impact on the environment and the nearby population (Aslam et al., 2022; Ali et al., 2023). This technique is particularly relevant for hazardous materials that require to be disposed in environmentally safe landfills (Abd-El Monsef and Smith, 2019).

## 5.4. Features extraction and classification

To reduce data dimensionality and focus detection models on the salient information, some works extract a combination of spatial and spectral features from the images.

**Detection** In Vambol et al. (2019), several brightness features computed from panchromatic images are used to detect waste dumps. In Parrilli et al. (2021), spectral information of multispectral images and spatial features extracted from panchromatic imagery are classified for detecting micro dumps and greenhouses. In Notarnicola et al. (2004), degraded areas are detected by recognizing characteristic spectral signatures that have a high likelihood of containing large-scale landfills. Spectral signatures are easily recognizable thanks to their temporal stability and can be used to identify recent landfills for which thermal anomalies and vegetation stress are undetectable. To increase detection performance, some researchers combine spectral signatures with textural spatial features (Salleh and Tsudagawa, 2002). The authors proved that the combination of spectral data (colour channels) and spatial information improves the accuracy.

## 5.5. Traditional Computer Vision techniques

Some works approached the solid waste detection task by applying CV techniques to multispectral and pansharpened images.

**Detection** A popular method used in RS image analysis is object-based classification, which pursues the identification and classification of objects or regions of interest rather than individual pixels. In Bilotta et al. (2012) and Didelija et al. (2022), multi-resolution segmentation is used to sequentially merge pixels and identify image objects, which are then classified as landfills. Similarly, Ulloa-Torrealba et al. (2023), define image objects using the SLIC algorithm while Chen et al. (2021) use a hierarchical segmentation method combined with a morphological index to detect construction waste from pansharpened images. In Faizi et al. (2020), municipal solid waste is detected using a pixel-based image classification model based on multispectral imagery.

## 5.6. Deep Learning Computer Vision techniques

Recent works adopted Convolutional Neural Networks (CNNs) trained on satellite

images to perform CV tasks such as image classification, object detection, and semantic segmentation. DL models learn relevant features of solid waste that can be exploited to detect landfills at scale.

**Detection** Two approaches detect aggregations of plastic waste by training a DL model to learn the spectral signature of plastic waste from multispectral images. In Kruse et al. (2023), pixel-level and patch-level neural network classifiers are trained on the spatial, spectral, and temporal dimensions of multispectral images. The pixel-level network extracts features across the spectral bands while the patch-level network classifies spatial features. In Lavender (2022), a pixel-level classifier integrates vegetation indices and multispectral channels to classify pixels as plastic waste or with a land use class. Devesa and Brust (2021) apply a semantic segmentation network to multispectral images where a model produces a binary mask denoting possible illegal large-scale landfills. Rajkumar et al. (2022) performs a similar segmentation task on pansharpened images to detect major landfills. Yang et al. (2022) and Yong et al. (2023) identify construction and demolition waste in urban areas. To cope with the complex spectrum and texture of such a class of waste, a model based on the DeepLabv3+ (Chen et al., 2018) architecture is used to perform semantic segmentation on high-resolution optical images. In Torres and Fraternali (2021), a binary scene classification model is trained on optical images at different ground resolutions. The binary classifier exploits ResNet (He et al., 2015) as the network backbone and augments it with a Feature Pyramid Network (FPN) (Lin et al., 2017) architecture. The FPN captures semantically strong features at all scales enabling the model to identify a wide variety of solid waste materials in urban and extra-urban scenarios. In the works Li et al. (2022), Zhou et al. (2023) and Sun et al. (2023), a CNN object detector is trained to localize waste dumps in urban scenarios. In Zhou et al. (2023) a modified YOLO (Redmon et al., 2016) network is used to localize industrial and household waste. Li et al. (2022) uses a Key Point Network to localize urban solid waste and Sun et al. (2023) predict the waste type (domestic, construction, covered, and agricultural waste).

**Monitoring** Yailymova et al. (2022) combines historical data with vegetation indices and LST computed from satellite multispectral data to analyze changes in landfills over time.

## 5.7. Other approaches

Some approaches identify and monitor landfills by combining geophysical or geological measurements and satellite data.

**Detection** In Di Fiore et al. (2017) the authors proved that electrical conductivity and induced polarization (a technique used to quantify the electrical chargeability of subsurface materials) are effective for the analysis of contaminated areas. Viezzoli et al. (2009) employ a special sensor mounted on a helicopter to detect anomalies in the subsurface electrical conductivity that may indicate the presence of a landfill.

**Monitoring** In Abou El-Magd et al. (2022), NDVI and LST indices are integrated with hydrological and geophysical measures collected during in-situ expeditions to perform an environmental assessment of a municipal landfill.

**Table 5** groups the techniques of the analyzed works by the type(s) of input data and by the category of data processing. A detailed description of the satellite missions and other data sources employed by each work is presented in Table 4.

From the survey of the adopted techniques, it can be noted that the approaches vary significantly based on the scale of the solid waste disposal to be detected. Small sites and large-scale landfills have very different characteristics that can be exploited for their identification. Large-scale landfills emit large quantities of gas that produce heat and thus thermal anomalies in the surface temperature and vegetation indices can be used as indicators for their detection. Smaller waste sites do not have a significant impact on thermal and vegetation indicators and therefore the analysis of optical and spectral signatures is preferable. In recent years, the adoption of DL for landfill detection has increased thanks to the development of powerful architectures that obtain very high performance even on aerial images. However, some recent works still obtain good performance with simpler techniques based on descriptive indices and with shallow Machine Learning models.

**Table 5**: *Techniques and input data (rows) grouped by waste management task (columns). The table reports references to the papers (defined in Table 1) that apply the various techniques to address different tasks.*

| Technique | Input data | Detection | | Monitoring | | Mapping | |
|---|---|---|---|---|---|---|---|
| | | Large-scale landfills | Urban waste dumps | Multi-temporal analysis | Environmental impact | Illegal dumping risk mapping | Landfill location assessment |
| Visual interpretation | Optical | (1) | (35) | | | | |
| | Pansharpened | | | (7) | | | |
| Descriptive indices extraction | Pansharpened | | (17) | | | | |
| | Multispectral | (4, 13, 23) | | (9, 11, 15) | (8, 19, 20, 28) | | |
| Multi-factor analysis | GIS data | | | | (27) | (14, 16, 21) | |
| | GIS and RS data | | | | | (5, 24, 37, 38) | (12, 22, 34, 44, 46) |
| Features extraction and classification | Optical | | (2) | | | | |
| | Panchromatic | (25) | (31) | | | | |
| | Multispectral | (3) | | | | | |
| Traditional CV techniques | Pansharpened | | (29) | | | | |
| | Multispectral | (10) | (26, 36, 48) | | | | |
| Deep Learning CV techniques | Optical | (49) | (32, 40, 47, 50) | | | | |
| | Pansharpened | (41) | (43) | | | | |
| | Multispectral | (30, 39, 45) | | (42) | | | |
| Other approaches | Geophysical data | (6, 18) | | | (33) | | |

# 6. Issues and research directions

Implementing solid waste detection models is challenging due to many variables such as diverse backgrounds, satellite image resolution, data availability, and reliable ground truth annotations. Some approaches have proved very effective for identifying waste disposal sites and illegal landfills. With the notable exceptions of Lavender (2022), Sun et al. (2023) and Rajkumar et al. (2022), the proposed solutions are tested in specific territories, with distinctive geographic characteristics, which questions the generalizability of waste detection models at a more global scale. Ultimately, the analysis of the current state-of-the-art demonstrates that there are still margins for improving the current level of maturity in this research field. The identified aspects highlight limitations related to the data availability, methodological issues in the development of the proposed approaches and gaps with respect to the requirements of practical waste management applications.

## 6.1. Open issues

### 6.1.1. Data related open issues

**Data set availability** Table 3 shows that very few data sets are available for training and evaluating solid waste detection models. The reasons may be the risk of exposing sensitive information about illegal activities, the adoption of a methodology that does not consider the repeatability of experiments, the terms of use of the images that disallow the publication of data, and the complexity and effort of producing data sets (Li et al., 2022). The lack of publicly available data could be a barrier for researchers interested in addressing waste management problems, because they would need to spend significant time and effort in creating a data set before starting to explore new methodologies.

**Coarse spatial resolution** Current generations of EO satellites offer sub-meter ground resolution, which however in many cases is still not sufficient to discriminate different waste materials. This limitation is even more critical in smaller municipal waste sites, where the presence of hazardous or polluting materials should require a higher urgency of intervention to reduce risks for the nearby population. The coarse ground resolution also prevents the accurate estimation of waste volumes, an important indicator of the severity of environmental impact.

### 6.1.2. Methodological and technical open issues

**Benchmarking** The lack of a widely adopted data set prevents the comparison of the different approaches, which in turn hinders the reliable evaluation of research progress. Inspiration can be taken from other CV and image analysis applications, which exist also in the RS field, such as aerial scene classification (e.g. AID (Xia et al., 2017), Million-AID (Long et al., 2021), NWPU-RESISC45 (Cheng et al., 2017)) and land cover characterization (e.g. UC Merced (Yang and Newsam, 2010), BigEarthNet (Sumbul et al., 2019), EuroSAT (Helber et al., 2019)). A benchmark should comprise a large and representative set of training, validation, and testing samples addressing diverse image resolutions and quality, object scales, scene configurations, solid waste materials, land uses, and geographical context. It should also specify a variety of downstream waste-related tasks, such as image-level binary and multi-label scene classification, and solid waste object detection or segmentation. Current datasets described in Section 4.2 have limitations that prevent them from becoming a global benchmark. Yin (2023) is annotated with only a few waste categories while Rajkumar et al. (2022) and Zhou et al. (2023) do not provide any material annotations, so these data sets cannot be used to recognize waste materials. Finally, Torres and Fraternali, 2023 provide a broad selection of waste classes but images only cover a limited geographic region, therefore models cannot properly generalize on global scenarios.

**Performance assessment procedures** Many works lack a quantitative assessment of the detection capability. A model is directly executed on a case scenario and the results are validated qualitatively, which makes it difficult to understand capability of the approach. The presented techniques are tailored to a specific use case, which cannot be easily generalized to other scenarios. This hinders the adoption of existing methods for new applications and the reuse of existing knowledge. Studies that report a quantitative evaluation use different performance metrics (accuracy, F1, AUC-ROC), which impedes the comparison of works. Often the data or the model is not published, preventing the reproducibility of the reported results. As a result, the performance of the proposed approaches under different metrics are not comparable because the evaluation cannot be reproduced exactly.

### 6.1.3. Application open issues

**Violation recognition and site risk assessment** RS analysis alone cannot tell legal from illicit waste disposal. An authorized waste landfill can act illegally by storing unauthorized materials or excessive quantities. A legal site, such a deposit of industrial materials, can visually appear as an illegal waste dump. No works report on methods to distinguish legal from illegal disposal, which is required for prioritizing interventions based on the level of risk.

## 6.2. Research directions

### 6.2.1. Data related research directions

**Creation of a global data set** RS image analysis has benefitted from such data sets as AID (Xia et al., 2017) for the classification of aerial scenes and UC-Merced (Yang and Newsam, 2010) for land use characterization. For solid waste detection, a large data set composed of images captured with multiple sensors (RGB, Panchromatic, Multispectral) could help standardize the comparison of models. An example in the domain of land use classification is the Million-AID data set (Long et al., 2021), a collection of one million Google Earth images annotated with 51 semantic categories. For the studies of waste sites evolution over time, the data set should comprise multiple snapshots of the same region at different times, sampled at short (e.g., revisit time) and mid-term (e.g., yearly) frequency.

**Future satellite constellations** The deployment of future satellites will increase the accuracy of state-of-the-art methods and fuel the research for novel approaches. An example is the *Albedo Satellite Constellation*[25] that will be deployed in Very Low Earth Orbit (VLEO) and will offer 10cm panchromatic imagery. The satellites will also be equipped with a 4-band multispectral sensor with 40cm spatial resolution and a LWIR thermal sensor with 2m GSD. The launch is scheduled for 2025 and the full constellation will offer 5 revisits per day. Such high resolutions were previously only achievable with expensive aerial surveys with longer revisit time. This constellation will allow researchers to develop techniques for more frequent monitoring of landfills and more accurate estimation of environmental impacts. New tasks could also be explored, such as accurately identifying a wider range of waste materials or estimating the volume of waste accumulations.

---

[25] https://www.satimagingcorp.com/satellite-sensors/albedo-10cm/

### 6.2.2. Methodological and technical research directions

**Waste material identification** Training detectors for identifying multiple waste types (e.g., the materials coded by the European Union (Eurostat, 2010)) enables better risk assessment and inspection management. EO satellites with spatial resolution up to 30cm allow models to discriminate waste classes such as rubble, tires, metal or wood. Dual image approaches, such as Alvarez-Vanhard et al. (2021), fuse EO and drone images and could help discriminate the diverse materials.

**Geospatial Foundation Models** Large pre-trained models, also known as *foundation models* (FMs), are trained on a large set of unlabeled data in a task-agnostic manner and can be fine-tuned to various downstream tasks. Thanks to the pre-training, fine-tuning an FM can outperform task-specific fully-supervised models while requiring less annotated data. In RS, NASA and IBM released a Geospatial AI Foundation Model[26] pre-trained on NASA's Harmonized Landsat Sentinel-2 (HLS) data set. The pre-trained backbone, named *Prithvi-100M* (Jakubik et al., 2023), could be exploited in the detection of solid waste for reducing the semantic gap due to the pre-training on collections of natural images for RS tasks. The effectiveness of RS pre-training is studied empirically in (Wang et al., 2023) and the authors find that the benefits are more evident for whole image tasks such as scene recognition and less apparent for pixel-level tasks such as image segmentation.

**Zero/Few Shot Learning for detecting rare waste types** In the surveyed data sets a few classes have a majority of labelled samples while others are scarcely represented. In the AerialWaste data set (Torres and Fraternali, 2023) the top 5 classes (rubble, bulky items, firewood, scraps, and plastic) account for 83% of the labels of 15 waste types. Suitable techniques should be devised for waste classes identification based on very few or even no prior visual examples. Learning with zero or few samples is addressed by zero- and few-shot approaches, which have demonstrated impressive results thanks to the advent of Visual Language Models (VLM) (Radford et al., 2021). A VLM exploits a massive amount of text, typically scraped from the Web, to supervise the training of a visual model through a simple ancillary task, such as image-to-text pairing. The extracted multimodal representations enable the use of knowledge mined from the text to classify images of previously unseen concepts. The application of VLM to RS tasks is in its infancy (Wen et al., 2023; Li et al., 2023) and many opportunities for improvement are open. The main difficulty of building a VLM for RS applications is that the amount of textual knowledge about RS concepts is much smaller than the text characterizing natural images. Furthermore, RS scenes have great structural and background diversity and weaker semantic correlations between entities observed in the context of RS, making the semantic reasoning at the base of zero-shot learning more difficult. This poses a challenge both for the construction of an effective VLM for RS applications (Li et al., 2023) and for the adaptation of such a model to the waste domain. However, if such a VLM could be built effectively, it would enable an array of very interesting applications, including the detection of previously unseen waste types, the automatic textual description of waste images, the grounding of image classifications by highlighting the relevant regions where the waste appears, and the implementation of query systems retrieving waste images from textual inputs.

---

[26] https://www.earthdata.nasa.gov/news/impact-ibm-hls-foundation-model

### 6.2.3. Application research directions

**Prioritization of sites** Waste detection via RS image analysis only captures information embedded in visual content and ignores valuable knowledge used by the environment and law enforcement agencies to estimate the risk and gravity of an illicit waste disposal and to decide for in-situ inspection. Such knowledge may be informal and dependent on many factors, such as local permissions and legislation, waste management objectives, investigation practices and budget constraints. A promising direction of research is the construction of a decision support system in which the expert knowledge and the output of a RS waste detection systems are integrated to produce a ranking of the candidate sites based on their risk level and on their priority for further investigation (Réjichi et al., 2015; Taha et al., 2023).

**Court-proof evidence collection** A major goal of waste detection is to identify illegal disposal and prosecute environmental crime. Such objective entails that the evidence collected meet the legal requirements for the use in court. The collection and processing of satellite images for the investigation and prosecution processes poses several changes, especially regarding privacy and veracity (Campbell, 2023; Nutter, 2018). Privacy concerns raise from the collection of very high-resolution images, which may infringe the right to privacy of individuals (Coffer, 2020). Veracity demands that the data collected via RS are trustworthy and demonstrably tamper-free. This is far from trivial because satellite images are the outcome of a multistep processing chain that transforms raw data into the products used for analysis. In addition to data trustworthiness, also the downstream analysis process that leads to an illegal site detection must meet admissibility criteria, such as the transparency and reproducibility of the machine-computed evidence. The use of automated waste detection tools in court poses formidable challenges, which demand the interdisciplinary collaboration of researchers from fields such as earth observation, sensor design, machine learning, data management, laws and social science.

## 7. Conclusions

This review presented the state-of-the-art for detecting and monitoring solid waste dumps in RS data with the aim of providing a practical analysis of the works in the field. The proposed approaches have been categorized, described, and compared and the publicly available data sets identified. The available RS data and the satellite products usable for solid waste detection and monitoring have been illustrated. The most significant open issues identified are mainly methodological and refer to the lack of a standard benchmark for assessing and comparing competing approaches, based on a large and global collection of RS images and well-defined inference tasks supported by agreed-upon performance evaluation metrics. Although establishing a reference dataset is a commendable initiative, the complexity of the problem should be considered. Effectively addressing the breadth of encountered scenarios requires the inclusion of a comprehensive set of samples characterized by heterogeneous geographical contexts and scene configurations capturing a wide selection of solid waste materials and landfill scale. Moreover, imagery from several EO sensors, represented at different ground resolutions, should be provided within the dataset. The identified research directions for the future are extremely exciting. The waste detection domain can benefit from the recent innovations in the CV arena, such as Vision Transformers, billion-scale foundation models and visual language models. However, such breakthroughs have impacted mostly the domain of natural image analysis and thus research opportunities are open for improving the state-of-the-art in both RS in general and solid waste detection research in particular.

## Declaration of Competing Interest

The authors declare that they have no known competing financial interests or personal relationships that could have appeared to influence the work reported in this paper.

## Acknowledgements

This work is supported by European Union's Horizon Europe project PERIVALLON - Protecting the EuRopean terrITory from organised enVironmentAl crime through inteLLigent threat detectiON tools, under grant agreement no. 101073952.

## CRediT authorship contribution statement

**Piero Fraternali:** Supervision, Conceptualization, Investigation, Writing - Review and Editing. **Luca Morandini:** Investigation, Writing - Original Draft, Writing - Review and Editing, Visualization. **Sergio Luis Herrera González:** Conceptualization, Investigation, Writing - Original Draft, Writing - Review and Editing.